\documentclass{article} 
\usepackage{nips15submit_e,times}
\usepackage{url}
\usepackage[numbers]{natbib}
\usepackage{comment}
\usepackage{url}
\usepackage{graphicx}
\usepackage{subcaption}
\usepackage{amssymb,amsmath,mathbbol}
\usepackage{rotating}
\usepackage[normalem]{ulem}
\title{Active Long Term Memory Networks}

\author{
Tommaso Furlanello\thanks{Corresponding Author} \\
University of Southern California\\
\texttt{furlanel@usc.edu} \\
\And
Jiaping Zhao \\
University of Southern California \\
\texttt{JustinZo@hotmail.com } \\
\AND
Andrew M. Saxe\\
Harvard University \\
\texttt{asaxe@fas.harvard.edu } \\
\AND
Laurent Itti\\
University of Southern California \\
\texttt{itti@usc.edu } \\
\And
Bosco S. Tjan\\
University of Southern California \\
\texttt{btjan@usc.edu} \\
}

\nipsfinalcopy 

\begin{document}

\maketitle
\begin{abstract}
Continual Learning in artificial neural networks suffers from interference and forgetting when different tasks are learned sequentially. This paper introduces the Active Long Term Memory Networks (A-LTM), a model of sequential multi-task deep learning that is able to maintain previously learned association between sensory input and behavioral output while acquiring knew knowledge. A-LTM exploits the non-convex nature of deep neural networks and actively maintains knowledge of previously learned, inactive tasks using a distillation loss \cite{hinton2015distilling}. Distortions of the learned input-output map are penalized but hidden layers are free to transverse towards new local optima that are more favorable for the multi-task objective. We re-frame the McClelland's seminal Hippocampal theory \cite{mcclelland1995there} with respect to Catastrophic Inference (CI) behavior exhibited by modern deep architectures trained with back-propagation and inhomogeneous sampling of latent factors across epochs. We present empirical results of non-trivial CI during continual learning in Deep Linear Networks trained on the same task, in Convolutional Neural Networks when the task shifts from predicting semantic to graphical factors and during domain adaptation from simple to complex environments. We present results of the A-LTM model's ability to maintain viewpoint recognition learned in the highly controlled iLab-20M \cite{Borji_2015_CVPR} dataset with 10 object categories and 88 camera viewpoints, while adapting to the unstructured domain of Imagenet \cite{russakovsky2015imagenet} with 1,000 object categories.

\end{abstract}
\section{Introduction}
The recent interest in bridging human and machine computations \cite{mnih2015human,liao2016bridging,lake2016building,kulkarni2016hierarchical} obliges to consider a learning framework that is continual, sequential in nature and potentially lifelong \cite{mikolov2015roadmap,tessler2016deep}.Therefore, such a learning framework is prone to interferences and forgetting.
On the positive side, intrinsic correlations between multiple tasks and datasets allow to train deep learning architectures that can make use of multiple data and supervision sources to achieve better generalization \cite{yosinski2014transferable}. The favorable effect of multi-task learning \cite{caruana1995learning,caruana1997multitask} depends on the shared parametrization of the individual functions and the simultaneous estimation and averaging of multiple losses. When trained simultaneously shared layers are obliged to learn a common representation, effectively cross-regularizing each task. Generally, in the sequential estimation case, the most recent task benefits from an inductive bias \cite{pratt1991direct,yosinski2014transferable} while older tasks become distorted \cite{mccloskey1989catastrophic, french1999catastrophic,goodfellow2013empirical} by unconstrained back-propagation \cite{hecht1989theory,le1990handwritten} of errors in shared parameters, a problem identified as Catastrophic Interferences (CI) between tasks \cite{mccloskey1989catastrophic}.

In humans the dyadic interaction between Hippocampus and Neocortex \cite{mcclelland1995there} is thought to mitigate the problem of CI by carefully balancing the sensitivity-stability \cite{hebb2005organization} trade-off such that new experiences can be integrated without exposing the system to risk of abrupt phase transitions. 

Think of the classical example of a child exploring new objects through structured play \cite{piaget2008psychology}, who samples from multiple points of view, lighting directions and generates movement of the object in space. This exploration creates inputs for the perceptual systems that span homogeneously all the underlying variation in viewing parameters and construct general purpose graphical, categorical or semantic level representations of its perception. Such representations can be used in the future to actively regularize experience in environments where exploration is constrained or costly.

We conjecture that CI arises because during the learning lifespan of a system the distribution of locally observed states \cite{crutchfield1999thermodynamic,shalizi2001causal} in the environment is non-stationary and potentially chaotic, while the neural representation of the environment with respect to its mode of variation as latent graphical \cite{kulkarni2015deep,bengio2013representation} and semantic \cite{haxby2001distributed,yamins2013hierarchical} factors must be stable and slowly evolving to store non-transient knowledge.

Early experiments on neural networks' ability to learn the meso-scale structure \cite{hinton1986learning} of their training environment required interleaved exposure to the different semantic variations. This heuristic is respected in modern architectures for object recognition \cite{krizhevsky2012imagenet} trained with stochastic mini batches with uniform, and possibly alternated, rich sampling of categories \cite{lecun2012efficient}. Analogously data augmentation regularizes the distribution of graphical factors. We hypothesize that the inability of CNNs to learn object categorization with strongly correlated batches is caused by interferences across the vast number of categories and latent graphical factors that must be memorized.

Similarly, the outstanding success of Deep-Q-Networks (DQN) \cite{mnih2015human,silver2016mastering} can partially be found in their intrinsic replay system \cite{lin1992self} that augments learning batches with state-action transitions that have not been observed in a long time. This procedure allows the creation of representations that DQNs transverse quasi-hierarchically \cite{zahavy2016graying} during play. It is difficult to imagine how these representations could be remembered if past states are not visited again through the replay system.

For the simplified case of Deep Linear Networks, it is possible to obtain an exact analytical treatment of the network's learning time as a function of input-output statistics and network depth \cite{saxe2013exact,saxelearning}. Suggesting that phase transitions typical of catastrophic interference might appear when the mixing time of the data generating process is longer than the learning time of the system, obliging the neural network to track the local dependencies between factors of variation instead of learning to represent the data generating process in the completeness of its ergodic state.

In this paper, we develop the Active Long Term Memory networks inspired by the Hippocampus-Neocortex duality and based on the knowledge distillation \cite{hinton2015distilling} framework. Our model is composed by a stable component containing the long term task memory and a flexible module that is initialized from the stable component and faces the new environment. We capitalize on the human infant metaphor and show that is possible to maintain the ability to predict the viewpoint of an object while adapting to a new domain with more images, object categories, and viewing conditions than the original training domain. Moreover, we discuss on the necessity to store and replay input from the old domain to fully maintain the original task.
\section{Related Work}
With a non-convex system to store knowledge in a changing environment, is important to understand what knowledge is synthesized in a neural network. The general intuition is that for hierarchical models with thousands of intermediate representations and millions of parameters it is difficult to identify the contained knowledge with respect to its parameters value across all layers. Without any guarantee of being in a unique global optimum, multiple configurations of weight parameters could sustain the same input-output map, making the association between parameters and knowledge vague.

The Knowledge Distillation (KD) framework \cite{buciluǎ2006model,hinton2015distilling} introduces the concept of model compression to transfer the knowledge of a computationally expensive ensemble into a single, easy to deploy, model using the prediction of the complex model as supervision for the compressed one. In the “born-again-tree” paradigm, Leo Breiman \cite{breiman1996born} proposed to use one tree model to predict outputs of random forest ensembles for better interpretability. In the KD framework, knowledge in neural networks is therefore identified in the input-output map without regard of its parametrization. Transferring knowledge between two neural architecture ("from teacher to student") therefore corresponds to supervised training of the student network using the logits of the original teacher network, or by matching the soft-probabilities that they induce.

 This framework received a lot of recent interest and has been extended to what is called Generalized Distillation \cite{lopez2015unifying} to incorporate some theoretical results of Vapnik's privileged information algorithm \cite{vapnik2015learning}. Recently large scale experiments \cite{ba2014deep,geras2015compressing,urban2016deep} have been carried on the ability to compress fully connected into shallow models or convolutional models into Long Short Term Memory \cite{hochreiter1997long} networks and vice-versa.
 
The introduction of double streams architecture inspired by the classical Siamese Network \cite{bromley1993signature} for metric learning, and the consequent generalization of the multi-task framework to domain adaptation \cite{long2015learning,rozantsev2016beyond} offers a natural extension of KD, where distillation happens between streams made to handle different data or supervision sources, but with a shared parametrization.
  
 The case for a strong effect of CI during sequential learning in deep neural networks has been shown, respectively between semantic \cite{goodfellow2013empirical} and graphical factors \cite{Borji_2015_CVPR}. In \cite{kulkarni2015deep} the authors are able to estimate an encoder-decoder model with correlated mini-batches using interleaved learning with carefully selected factors ratio and ad-hoc clamping of the neurons learning rate. In \cite{rusu2015policy,parisotto2015actor} multiple Atari games are learned with interleaved distillation across games, the correct ratio between batch size and interleaving was again carefully curated and crucial for the algorithm success. The authors \cite{rusu2015policy} also present a novel self-distillation framework remembering the Minskian Sequence of Teaching Selves \cite{minsky1988society}.
 
 \section{The A-LTM Model}
We approach the problem of learning with a sequence of input-outputs that exhibits transitions in its latent factors using a dual system. Our model is inspired from the seminal theory of McClelland on the dyadic role of hippocampus and neocortex in preserving long-term memory and avoiding catastrophic interferences in mammals \cite{mcclelland1995there}.

The first A-LTM component is a mature and stable network, \textit{Neocortex} ($N$), which is trained during a development phase in a homogeneous environment rich of supervision sources. To prevent the interference of new experience with previously stored functions, the $N$ networks's learning rate is drastically reduced during post-developmental phases in an imitation of the visuo-cortical critical period of plasticity \cite{wiesel1963effects}.

The second component in the A-LTM is a flexible network, \textit{Hippocampus} ($H$), which is subject to a general unstructured environment. $H$ weights are initialized from $N$ and learning dynamics are actively regularized from $H$ output activity. 

This dual mechanism allows to maintain stability in $N$ without ignoring new inputs. $H$ can quickly adapt to new information in a non-stationary environment without generating a risk for the integrity of knowledge already stored in $N$. Furthermore long term information in $N$ are actively distilled into $H$, with the effect of constraining the gradient descent dynamics ongoing in $H$ towards a better local minimum able to sustain both new and old knowledge. Operationally:

\begin{enumerate}
	\item During development: $N$ is trained in a controlled environment where multiple examples of the same object and of its potential graphical transformation are present. We train $N$ with a multi-task objective to predict both semantic (category) and graphical (camera-viewpoint) labels of the object. After convergence, the learning rate of $N$ is set to 0.
	\item During maturity: $H$ is initialized from $N$ and faces a novel environment, where objects are typically available from a single perspective and the number of categories is increased by two orders of magnitude (from 10 to 1000 classes). $H$ is trained with a multi-task objective to predict both the new higher dimensional semantic task and the output of $N$ with respect to the developmental tasks, in this case $N$'s ability to differentiate between different point of view of the same object.
\end{enumerate}
A-LTM networks relies on the core idea that all the tasks to memorize need simultaneous experience to find a multi-task optimum that is a critical point for all the objectives. Therefore, if the environment has unstable input-output because of missing labels, $H$ has to rely on predictions from $N$. In the complementary case, where instabilities are generated by changes in the distribution of input, an auxiliary replay mechanism is also necessary.

\textbf{A Sequential Environment}

We study the situation where an agent interacts with a sequential environment, defined by the joint distribution $P(y,x)$ of visual stimuli $x\in X$ and their binary latent factors $y\in Y$. The agent receives information through a perceptual mechanism $\Psi(x): X \mapsto S$ and makes decision based on a hierarchical representation of its percept $\phi (s): S \mapsto \Phi^{d}$. The agent's goal is to name the underlying latent factors for each stimulus. Actions are chosen by the agent with a n-way soft-max that transforms $\phi^{d} (s)$, the last layer of the hierarchical representation of sensory inputs, into a probability distribution over actions. 

The environment responds to actions with a supervised signal $y\in Y$ informing the agent on the correct action given a particular stimulus. The hierarchical representation $\phi$ is updated in order to minimize the cross-entropy loss $\mathcal{L}(\phi^{d} (s), y)$ .
This task is computationally tedious because of the vast range of possible transformations in sensory inputs $\tau: S \mapsto S$ that do not have any meaningful consequence on the semantic nature of the stimulus. We call these transformations latent graphical factors. Perceptual inputs to our system $s$ have therefore two modes of variation: 
\begin{enumerate}
	\item Variations in semantic factors that alters the category of stimulus $x$ and its percept $s$, parametrized by the subcomponent $y_{s}\in y$.
	\item Variations in graphical factors that are invariant with respect to the category of $x$ but alter its percept $s$, parametrized by the subcomponent $y_{g}\in y$.
\end{enumerate}
Catastrophic interference happens when the distribution of supervised signals $P(y,x)$ is not homogeneous. While modeling environment's 
non-stationarity in the language of stochastic process could lead to interesting insights, we limit ourselves to the simple regime with a single discrete transition from $P^{1}(y^{1},x^{1}) $ to $P^{2}(y^{2},x^{2})$.

\textbf{Bridging Sequential and Multi-Task Learning via Knowledge Distillation}

Let the multitask function representing the input - output maps of network be $ f(w_{0},w_{1},w_{2};x): X \mapsto Y$, where $w_{0}$ are shared parameters and $w_{1},w_{2}$ the task-specific parameters defining a map from the common representation to the individual tasks $y^{1},y^{2}$.

Sequential Learning corresponds to solving in this sequence the following two optimization problems. First the minimization of the cross entropy loss $\mathcal{L}(f(w_0,w_1;x_{1}),y_{1}) $ between the environment data generating process $P^{1}(y^{1}|x^{1}) $ and the softmax probability distribution induced over the task 1 predictions $f(w_0,w_1;x_{1})$, with a Gaussian initializations $ w^{0}_0$ and $ w^{0}_1$:
\begin{equation}\label{min:se1}
\begin{aligned}
& \underset{w_0,w_1}{\text{min}}
& & \mathcal{L}(f(w_0,w_1;x_{1}),y_{1}) \\
& \text{s. t.}
& & w^{0}_0 \sim \mathcal{N}(0,\sigma)\\
&& & w^{0}_1 \sim \mathcal{N}(0,\sigma)
\end{aligned}
\end{equation}
with solutions $w^{*}_0$, $w^{*}_1$. Followed by analogous problem for the second environment with starting condition equal to the solution of the previous problem for the shared parameters $w_0$:
\begin{equation}\label{min:se2}
\begin{aligned}
& \underset{w_0,w_2}{\text{min}}
& & \mathcal{L}(f(w_0,w_2;x_{2}),y_{2}) \\
& \text{s. t.}
& & w^{0}_0 = w^{*}_0\\
&& & w^{0}_2 \sim \mathcal{N}(0,\sigma)
\end{aligned}
\end{equation}
with solutions $w^{**}_0$, $w^{**}_2$, making $f(w^{**}_0,w^{*}_1;x_{1})$ an unlikely critical point of problem \ref{min:se1}.\\
Multitask learning mitigates the problem of CI by averaging weight updates across different objectives and corresponds to solving \ref{min:se1} and \ref{min:se2} simultaneously, with an omitted scaling factors between the two objectives:
\begin{equation}\label{min:mt1}
\begin{aligned}
& \underset{w_0,w_1,w_2}{\text{min}}
& & \mathcal{L}(f(w_0,w_1;x_{1}),y_{1})+\mathcal{L}(f(w_0,w_2;x_{2}),y_{2}) \\
& \text{s. t.}
& & w^{0}_0 \sim \mathcal{N}(0,\sigma)\\
&& & w^{0}_1 \sim \mathcal{N}(0,\sigma)\\
&&& w^{0}_2 \sim \mathcal{N}(0,\sigma)\\
\end{aligned}
\end{equation}
 The drawback of this approach is that $(y^{1},x^{1};y^{2},x^{2})$ must be available to the network during the whole training phase.
 
 In absence of labels $y_1$ for the old task, Knowledge Distillation can be used as a surrogate. In A-LTM the stable module $N$ trained with the problem \ref{min:se1} can be used to hallucinate missing labels. This way, the new learning phase of $H$ can be recast in the multi-task framework even in absence of external supervision solving the following:
 \begin{equation}\label{min:altm1}
 \begin{aligned}
 & \underset{w_0,w_1,w_2}{\text{min}}
 & & \mathcal{L}(f(w_0,w_1;x_{1}),f(w^{*}_0,w^{*}_1;x_{1}))+\mathcal{L}(f(w_0,w_2;x_{2}),y_{2}) \\
 & \text{s. t.}
 & & w^{0}_0 = w^{*}_0\\
 && & w^{0}_1 = w^{*}_1\\
 &&& w^{0}_2 \sim \mathcal{N}(0,\sigma)\\
 \end{aligned}
 \end{equation}

 With respect to the availability of inputs, if they belong to the same modality and share the same distribution of graphical factors of variation, the active stream of perception can be used to train both active and inactive tasks in the flexible module. Otherwise, either inputs have to be stored or the networks must rely on some generative mechanism to generate imaginary samples for the non-ongoing task. In eq \ref{min:altm1} we assume the presence of a replay mechanism allowing the A-LTM networks to have access to both $x_1$ and $x_2$ during training. We relax this assumption in the experiments and present results also for the problem with objective function $\mathcal{L}(f(w_0,w_1;x_{2}),f(w^{*}_0,w^{*}_1;x_{2}))+\mathcal{L}(f(w_0,w_2;x_{2}),y_{2})$.
 
 \textbf{Beyond Active Maintenance, Memory Consolidation}\\
 While in this article we focus on the early phase of memory maintenance, a successive phase called memory consolidation, where knowledge is distilled from $H$ to $N$ is necessary for non-active long term memory.
 
 We interpret the phase of learning that we model as the step right before consolidation. An irreparable damage to $H$ would create a memory loss equivalent to temporally graded retrograde amnesia. In case of bilateral lesions \cite{scoville1957loss} to the hippocampus previously known information stored in $N$ would be safe, but recent adaptations to newly known environments only stored in $H$ would be completely lost.
 
\section{Experiments}
As a first illustration of the Active Long Term Memory framework, we conduct a sequence of experiments on sequential, multitask and A-LTM learning. We employ three datasets in experiments of increasing complexity. We first show that interferences emerge in the trivial case sequential learning of the same function in deep linear networks. We analyze the situation where complete forgetting happens during sequential learning of semantic and graphical factors of variation. Finally, we demonstrate the ability of A-LTM to preserve memory during domain adaptation with and without access to replays of the previous environment. The datasets used are:
\begin{itemize}
	\item \textbf{Synthetic Hierarchical Features} \cite{saxelearning}: we employ a procedure based on independent sampling from a branching diffusion process, previously used to study Deep Linear Networks learning (DLNs) dynamics, to generate a synthetic dataset with a hierarchical structure similar to the taxonomy of categories typical to the domain of living things.
	\item \textbf{iLab20M} \cite{Borji_2015_CVPR}: in this highly controlled dataset, images of toy-vehicles are collected over a turntable with multiple cameras, with different lighting directions. Ilab totals 704 objects from 15 categories, 8 rotating angles, 11 cameras 5 lighting conditions and multiple focus and background, for a complete dataset of over 22M images. Each object has 1320 images per background. Because of the highly controlled nature of the dataset, we reach high accuracy with a random subset of ~850k images from 10 categories and 11*8 camera positions: thus, we do not exploit variations in all the other factors for a matter of simplicity. iLab20M is freely available.
	\item \textbf{Imagenet 2010} \cite{russakovsky2015imagenet}: is an image database with 1000 categories and over 1.3 M images for training set. Imagenet is the most used dataset for large scale object recognition, since Alexnet \cite{krizhevsky2012imagenet} won the 2012 submission only deep learning model have been on the leaderboard of the annual ILSVRC competition. 
\end{itemize}
Except for the DLNs experiments, we train using stochastic gradient descent Alexnet's like Architecture with drop-out without any form of data augmentation on a central 256-by-256 crop of the images. We train until convergence,1 to 3 epochs, in iLab20M and for 10 epochs in Imagenet. During multi-domain and A-LTM, the loss function associated to iLAB is scaled by a factor of 0.1, as larger factors tend to stuck the Imagenet training in local minima. Knowledge distillation is implemented using an $l_2$ loss between the decision layers of the N and H network of both iLab categories and viewpoints; we adopted the $l_2$ instead of cross-entropy to avoid tuning an additional parameter for the teacher soft-max temperature. All of the experiments are implemented in Matconvnet \cite{vedaldi2015matconvnet} and run three Nvidia GPUs, two Titan X and one Tesla K-40.
\subsection{Catastrophic Interference in Deep Linear Networks}

We first study the aforementioned discrete transition in Deep Linear Networks \cite{saxe2013exact} for the trivial case of learning sequentially the same function. DLNs exhibits interesting non-linear dynamics typical of their non-linear counterpart, but are amenable to interpretation. Their optimization problem is non-convex and, since their deep structure can be transformed to a shallow linear map by simple factorization of layers, each deep local minimum can be compared to the shallow global optima.

In our experiments, the network architecture has an input layer, a single hidden layer, and two output layers, one for each task. We examine the trivial situation where Task A is equivalent to Task B. In this situation, obviously a multi-task solution corresponds to the network re-learning the hidden to output layer without any modification of the input to hidden layer. Alternatively, multiple solutions that maintain the input-output relationship constant but modify weights of the input-hidden layer, therefore requiring and adjustment of the hidden-output weights, are possible.
\begin{figure}
	\begin{subfigure}{.5\textwidth}
		\centering
		\includegraphics[width=.8\linewidth]{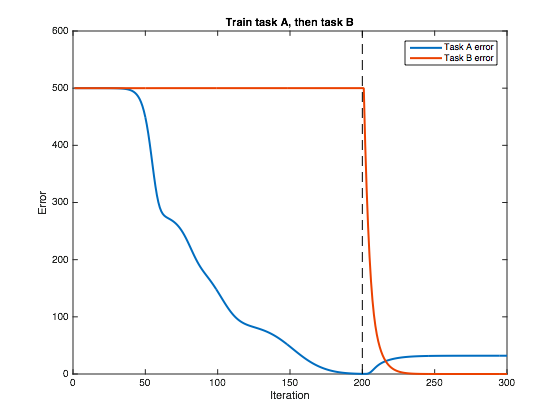}
		\caption{Sequential Learning: Task A $\mapsto$ B}
		\label{fig:sfig1}
	\end{subfigure}%
	\begin{subfigure}{.5\textwidth}
		\centering
		\includegraphics[width=.8\linewidth]{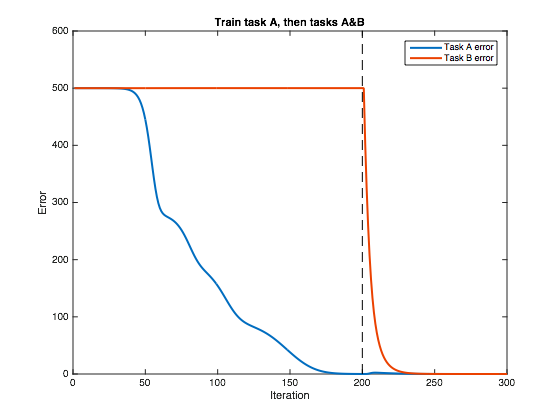}
		\caption{Multitask Learning: Task A $\mapsto$ A+B}
		\label{fig:sfig2}
	\end{subfigure}
	\begin{subfigure}{.5\textwidth}
		\centering
		\includegraphics[width=.8\linewidth]{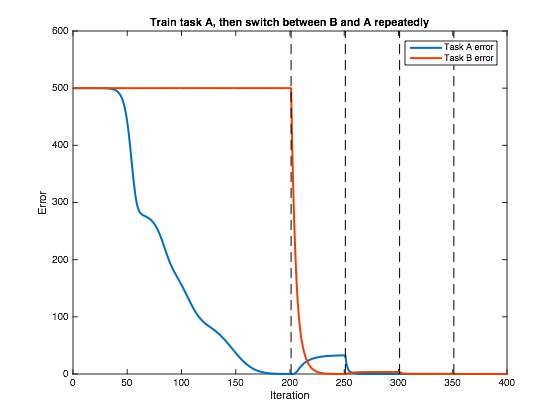}
		\caption{Interleaved Learning: Task A $\mapsto$ B $\mapsto$ A $\mapsto$ B}
		\label{fig:sfig3}
	\end{subfigure}
	\begin{subfigure}{.5\textwidth}
		\centering
		\includegraphics[width=.8\linewidth]{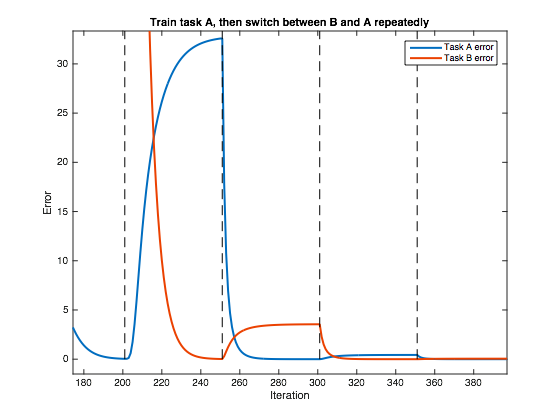}
		\caption{Zoom of (c): Catastrophic Interference}
		\label{fig:sfig4}
	\end{subfigure}
	\caption{Catastrofic Intereference experiments with Deep Linear Networks: Task A  is equivalent to Task B}
	\label{fig:lincat}
\end{figure}

In Fig \ref{fig:sfig1}, we first train the network with respect to Task A till convergence and then begin training with respect to the identical Task B; we find that back-propagation systematically has an effect on the input-hidden layer, generating interference of the original network. In Fig \ref{fig:sfig2}, multi-task learning does not have the same problem. In Fig \ref{fig:sfig3}-\ref{fig:sfig4} we instead alternate task every 50 epochs, generating reciprocal interference of decreasing intensity that delays convergence for more than 100 epochs with respect to the multi-task case.

\subsection{Sequential and Multi-task Learning of Orthogonal Factors on iLab20M}
Because of its parametrized nature, iLab20M is a good environment for the early training phase of models that can be taught to incorporate the ability to predict multiple semantic (like categories and identities) and graphical factors (like luminance or viewpoint).

 We train single task architectures to classify either between 10 Categories (ST-Category ) or 88 Viewpoints (ST-Viewpoint) and use its weights at convergence to warm up a network for the complementary task. We compare the Inductive Bias and Catastrophic Interference effect to single and multi-task solutions.

		\begin{figure}

						\label{combo:ilab}

			\begin{subfigure}{0.4\textwidth}
				\centering
					\includegraphics[width=0.8\linewidth]{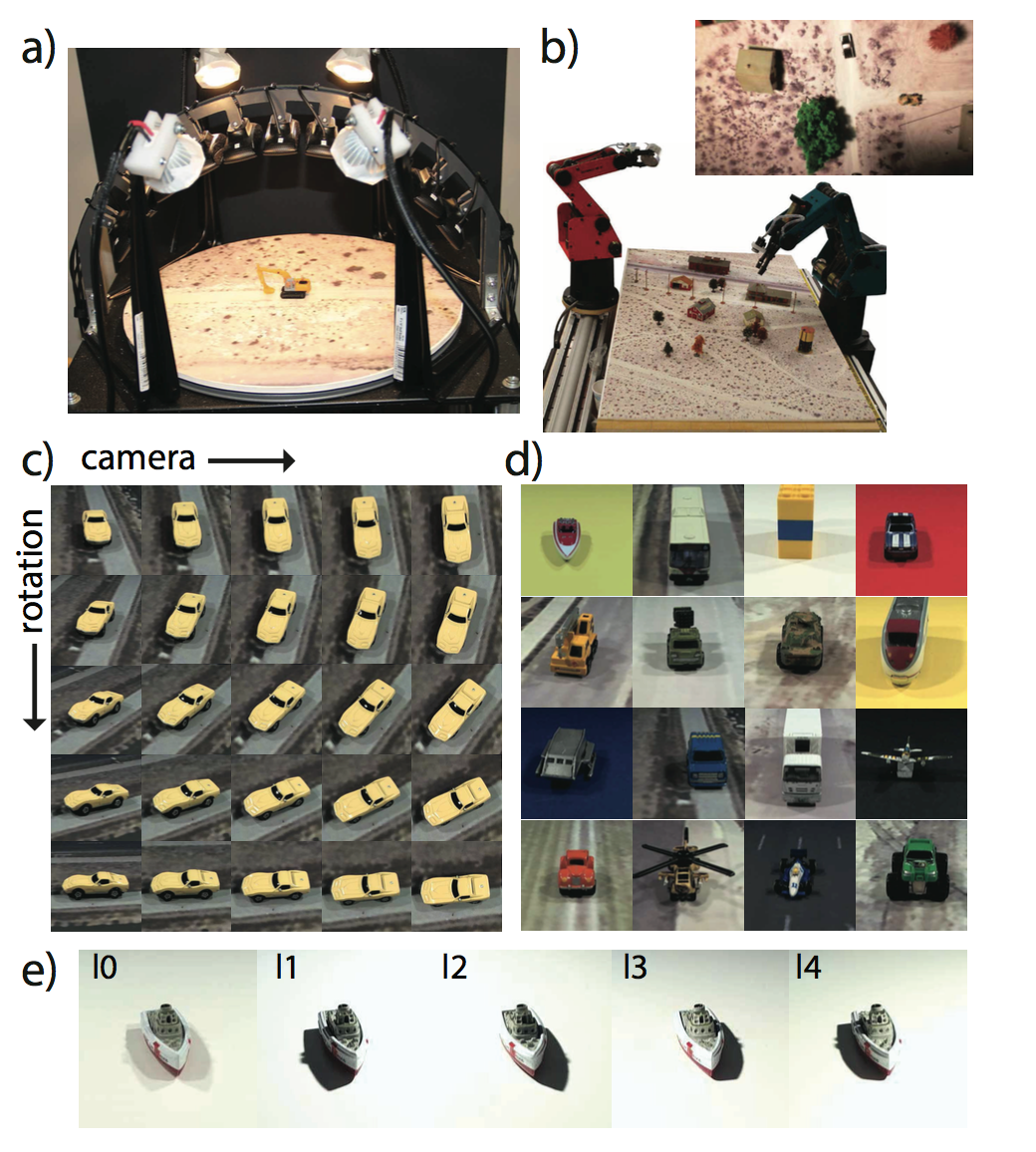}

					\caption{Turntable setup for iLab20M, figure from \cite{Borji_2015_CVPR} } 
				
					\label{fig:Ilab}
			\end{subfigure}
			\begin{subfigure}{0.6\textwidth}
					\centering

\label{table:ilab}

					\begin{tabular}{lll}
						& Categories                 & Viewpoints       \\ \hline
						\multicolumn{1}{l|}{ST - Category}   & \multicolumn{1}{l|}{0.81}          &     /       \\ \hline
						\multicolumn{1}{l|}{ST - Viewpoint}   & \multicolumn{1}{l|}{/}            & \textit{0.94}     \\ \hline
						\multicolumn{1}{l|}{Cat $\mapsto$ View} & \multicolumn{1}{l|}{\textit{\textbf{0.12}}} & 0.93          \\ \hline
						\multicolumn{1}{l|}{View $\mapsto$ Cat} & \multicolumn{1}{l|}{0.84}          & \textit{\textbf{0.02}} \\ \hline
						\multicolumn{1}{l|}{MT - Cat/View}   & \multicolumn{1}{l|}{\textit{0.85}}     & 0.91          \\ \hline
					\end{tabular}
\caption{\textbf{Test set Accuracy for experiment 4.2: }ST for single task model, $\mapsto$ indicate sequential learning, MT for multitask learning. The effects of Catastrophic Interference in bold.}
						
			\end{subfigure}

								\caption{\textbf{a):} example of turntable setup and of multiple-camera viewpoint, lighting directions and objects from iLab20M. \textbf{b):}: Results of Sequential and Multi-Task learning of \textit{categories} and \textit{viewpoitns} onf iLab20M}
		\end{figure}

The results in table \textit{b} of fig 2 confirm the inductive bias of Viewpoints over Category and the dramatic interference of sequential training between the two tasks. 
Finally, as expected, multi-task learning can easily learn both tasks, incorporating the inductive bias from viewpoints to categories.

\subsection{Sequential, Multi-Task and A-LTM Domain Adaptation over Imagenet}
Imagenet is a popular benchmark for state of the art architecture for visual recognition because of its 1000 categories and over 1M images. We compare the adaptation performance to Imagenet of multiple architectures: a single task network with Gaussian initialization (ST - Gauss), sequential transfer from the iLab20M multi task network (ST-iLab), multi-domain networks trained simultaneously over the iLab and Imagenet tasks either initialized randomly (MD - Gauss) or from the iLab20M multi task network (MD - iLab).

As seen in Table \ref{tab:imag} we confirm the general intuition developed in the previous experiment. Initializing from iLab has a positive effect on the transferred performance; moreover unconstrained adaptation completely wipes the ability of the network to perform viewpoint identification. Multi-Task/Domain learning is able to reach the same performances on Imagenet while maintaining almost completely the viewpoint detection task.

\textbf{A-LTM: Domain Adaptation from iLab20M to Imagenet without extrinsic supervision}\\
While Multi-task learning seems to be a good strategy for finding local minimum able to mantain multiple tasks it requires two expensive sources of supervision: Images and Labels from the original domain.

The A-LTM architecture is able to exploit the absence of supervision by using its stable component to hallucinate the missing labels and convert the otherwise sequential learning in a multi-task scenario. If the distribution of input is homogeneous across datasets, i.e $P(y_1|x_1)=P(y_1|x_2)$ the input-output map of the stable network can be expressed, therefore distilled, using only images from the new domain as input avoiding completely the limitations of multi-task learning (A-LTM - naive). In the contrary case a replay system either based on re-generation or storage of the past input is necessary (A-LTM - replay). 

The multi-task architecture of experiment 4.2 is used as stable component $N$ for both experiments, i.e. we use it to for weight initialization of $H$ and as a source of supervision for KD in both A-LTM models. 

\begin{figure}
	\centering
	\includegraphics[width=1\linewidth]{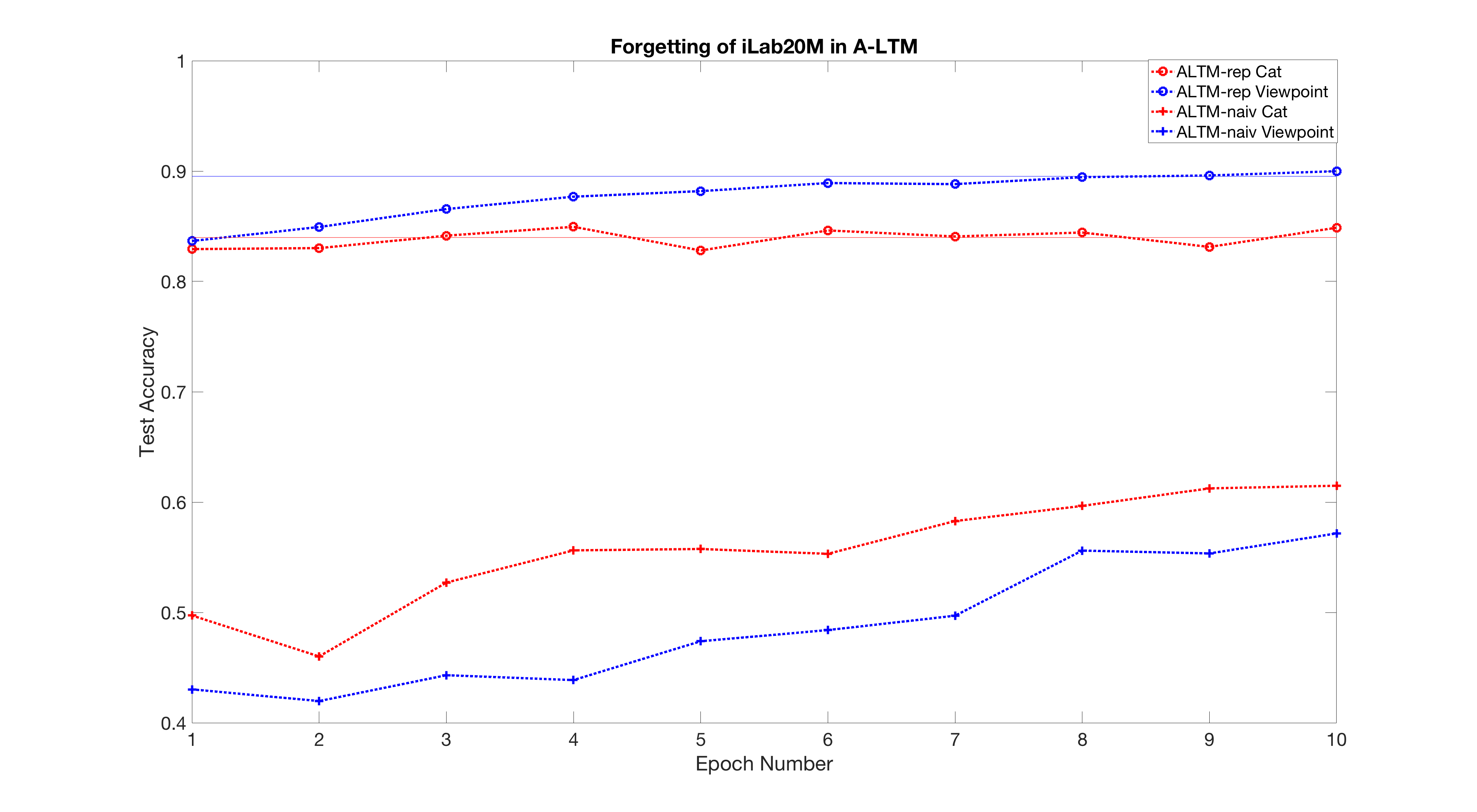}
	\caption{\textbf{Test Set Accuracy on iLab20M of A-LTM:} categories in blue and viewpoints in red. A-LTM with replay is indicated by circles while A-LTM without replay with plus signs, the dashed horizontal lines represents the accuracies at initialization. Both models suffers from an initial drop in accuracy during the first epochs but the A-LTM network that has access to samples of the original dataset is able to fully recover.}
	\label{fig:altm}
\end{figure}

The results on Imagenet in table \ref{tab:imag} show that the A-LTM architectures are able to maintain the long term memory of the Viewpoint task at the cost of a slower adaptation. The reduced memory performance in table \ref{tab:imag} of A-LTM-naive (without replay), especially the strong initial drop, for both viewpoints and categories as illustrated in figure \ref{fig:altm} is indicative of the strong shift in underlying factors between the two datasets and the importance of generative mechanisms to re-balance these differences with replay.

\begin{table}
	\centering
	\caption{Test set Accuracy of domain adaptation over Imagenet and Memory of the Viewpoint task for iLAB for single task (ST), multi-task/domain (MD) and Active Long Term Memory networks with and without replay.}
	\label{tab:imag}
	\begin{tabular}{lll}
		& Imagenet              & Viewpoints       \\ \hline
		\multicolumn{1}{l|}{ST - Gauss Init} & \multicolumn{1}{l|}{0.44}     &            \\ \hline
		\multicolumn{1}{l|}{ST - iLab Init} & \multicolumn{1}{l|}{0.46}     & \textit{\textbf{0.03}} \\ \hline
		\multicolumn{1}{l|}{MD - Gauss Init} & \multicolumn{1}{l|}{\textit{0.44}} & 0.81          \\ \hline
		\multicolumn{1}{l|}{MD - iLab Init} & \multicolumn{1}{l|}{0.45}     & \textit{0.84}     \\ \hline
		\multicolumn{1}{l|}{A-LTM - naive}  & \multicolumn{1}{l|}{\textit{0.40}} & \textbf{0.57}     \\ \hline
		A-LTM - replay            & 0.41                & 0.90         
	\end{tabular}
\end{table}

\section{Discussion}
In this work we introduced a model of long term memory inspired from neuroscience and based on the knowledge distillation framework that bridges sequential learning with multi-task learning. We show empirically that the ability to recognize different viewpoints of an object can be maintained also after the exposure to millions of new examples without extrinsic supervision using knowledge distillation and a replay mechanism. Furthermore we report encouraging results on the ability of A-LTM to maintain knowledge only relying on the current perceptual stream.
The theoretical analysis of DLNs linking convergence time of stochastic gradient descent to input-output statistics and network depth, and the plethora of tricks developed to successfully train deep networks suggest a potential relationship with the vanishing gradient problem \cite{hochreiter1997long,hochreiter2001gradient}. In order to learn in complex environments whose data generating process take a long time to mix it is necessary to use deeper architectures that have a longer convergence time, that in turn are plagued by the problem of propagating the gradient across multiple layers. 

While it seems easy to confuse the two memory problems, in supervised classification with Long Short Term Memory networks a careful balance of positive and negative examples in the mini-batches is crucial to the model performance, similarly to the interleaved scheme for sampling categories in CNNs.


\end{document}